\documentclass[final,3p,times,twocolumn]{elsarticle}

\usepackage{amsmath}
\usepackage{amssymb}
\usepackage{enumerate}
\usepackage{graphicx,url}
\usepackage{epsfig}
\usepackage{amsmath}
\usepackage{hhline}
\usepackage{multirow}
\usepackage{float}
\usepackage{hyperref}
\usepackage{enumerate}
\usepackage{graphicx,url}
\usepackage{graphicx}
\usepackage{xcolor}

\begin{document}

\begin{frontmatter}

\title{Real-Time Trajectory Optimization in Robot-Assisted Exercise and Rehabilitation}

\author[First]{Humberto De las Casas} 
\author[Second]{Nicholas Chambers} 
\author[First]{Hanz Richter}
\author[Second]{Kenneth Sparks}

\address[First]{Mechanical Engineering Department, Cleveland State University, Cleveland, OH 44115 USA.} 
\address[Second]{Health and Human Performance Department, Cleveland State University, Cleveland, OH 44115 USA.} 

\begin{abstract}
{This work focuses on the optimization of the training trajectory orientation using a robot as an advanced exercise machine (AEM) and muscle activations as biofeedback. Muscle recruitment patterns depend on trajectory parameters of the AEMs and correlate with the efficiency of exercise. Thus, improvements to training efficiency may be achieved by optimizing these parameters. The optimal regulation of these parameters is challenging because of the complexity of the physiological dynamics from person to person as a result of the unique physical features such as musculoskeletal distribution. Furthermore, these effects can vary due to fatigue, body temperature, and other physiological factors. In this paper, a model-free optimization method using Extremum Seeking Control (ESC) as a real-time optimizer is proposed. After selecting a muscle objective, this method seeks for the optimal combination of parameters using the muscle activations as biofeedback. The muscle objective can be selected by a therapist to emphasize or de-emphasize certain muscle groups. The feasibility of this method has been proven for the automatic regulation of an ellipsoidal curve orientation, suggesting the existence of two local optimal orientations. This methodology can also be applied to other parameter regulations using a different physiological effects such as oxygen consumption and heart rate as biofeedback.}
\end{abstract}

\end{frontmatter}
\noindent \textbf{Number of words in the document:} 3352.
\section{Introduction}
\noindent The performance of human exercise depends on the training parameters of the exercise machine. This performance is strictly important to increase efficiency and reduce the probabilities of injuries. Only in 2019, 468,315 injuries were reported in the US \citep{Gym_Injuries_0}. It is estimated that more than 2 of 5 gym users have had at least one injury while working out \citep{Gym_Injuries_1}. Therefore, a smart system for the automatic regulation of these parameters is worth developing. In this work, we explore the feasibility of this approach.\\

\noindent Research in areas of human-machine interaction (HMI) such as human performance and rehabilitation has received a lot of attention in recent years. For instance, performance enhancements have been evidenced by using robots as advanced exercise machines (AEMs). AEMs have proven to be highly valuable to human performance and rehabilitation developments over the past years by combining exercise physiology with technology \cite{MOTOR}. AEMs have electric motors and control systems able to produce controllable trajectories and impedances even in microgravity. As a result, the applications of these devices can prove to be beneficial for astronauts by allowing them to receive similar stimulus as to that achieved under the Earth's gravity forces \cite{ECC,ECC2,NASA_2,NASA}. The ability of the AEM to provide configurable trajectories and impedances makes it possible to enhance training patterns with a higher versatility than conventional training methods \cite{EXP_SET_5,EXP_SET_2,DELASCASAS1}. Previous researches using a powered rowing machine \cite{HUM_THE} and 4 DOFs robot reported diversity on the musculoskeletal, cardiovascular and cardiorespiratory systems through the manual regulation of trajectory and impedance parameters \cite{DELASCASAS2,TP_PENDING}. HMI applications focusing on rehabilitation have also achieved performance improvements using robotics to help people with disabilities to recover some of the motor skills that they used to have \cite{Shoulder_Rehab_Robot,MOT2,MOT3}.\\

\noindent The integration between information technology and robotics has been successfully applied in several interactive environments to support the improvement of human exercise and rehabilitation practices \citep{EXP_SET_2,EXP_SET_1}. The objective of this technology is to enhance manual dexterity similarly to that seen in virtual reality games \citep{EXP_SET_3}. One of the first robotic systems developed for therapy was the MIT-Manus (MITM). This robot-aid therapy system has assisted stroke patients by helping them to improve their motor skills for over 30 years now. MITM works by fitting the arm into a brace attached to the robot arm. Meanwhile, the subject is required to follow a target trajectory. This activity effectively assists the person in creating new neural connections which, over time, enable the re-learning of certain muscular processes \citep{EXP_SET_4II,EXP_SET_4I}. Several extensions and new developments for MITM \citep{EXP_SET_5} have been performed using manual regulation of the system parameters.\\

\noindent The research regarding robot-aid systems also goes into exercise training capabilities using integrated smart capabilities. The integration of HMI systems with smart controllers is known as Smart-HMI \cite{smart_HMI}. This novel mechatronic discipline considers the human dynamics inside of the closed-loop system. Thus an opportunity to unite human physiological understanding with mechanical, electrical, and information technology is presented. The first model-free training optimization has been reported using a single degree of freedom (DOF) manipulator with variable speed as the AEM to optimize the user's power output \cite{Perry_Horowitz_1,Perry_Horowitz_2}. The optimization was performed using the biomechanical configuration of the user which was real-time estimated with its force-velocity relationship. The previous research was replicated using a similar 1-DOF AEM and 2 Extremum Seeking Control (ESC) algorithms \cite{ZhanES}. These algorithms working independently were used to maximize the user's power output and estimate the user's torque. Regarding other research into the area of rehabilitation, biological model estimations have been successfully developed and reported \cite{adapt_rehab_game,smart_rehab}. The semi-automatic selection of machine parameters based on biofeedback has made it possible to achieve better results when compared to conventional rehabilitation practices.\\ 

\noindent Typical physiological measurements collected for human performance research include electromyography (EMG), heart rate (HR), oxygen consumption (VO$_2$), and energy requirements. Despite the simplicity of their measurements, physiological effects are complex as everyone has varying physiological functioning. They depend on the unique features of each person (organismic variables) such as force capacity, musculoskeletal distribution, body mass index, and flexibility. Besides, they display time-varying dynamics due to fatigue, body temperature, level of hydration, etc. As a result, modeling and controlling muscle performance is challenging but necessary in some areas of research \cite{Sim_J1}. Human dynamic models relating physical activity with muscle activations have been reported \cite{holly_phdthesis}. However, these models do not take into consideration the time-varying effects of training. They are only based on the dynamics of the musculoskeletal distribution. Consequently, they are useful and reliable for simulations, but not recommended for real-time experiments.\\

\begin{figure}[ht]
\begin{center}
\includegraphics[width=7.5cm]{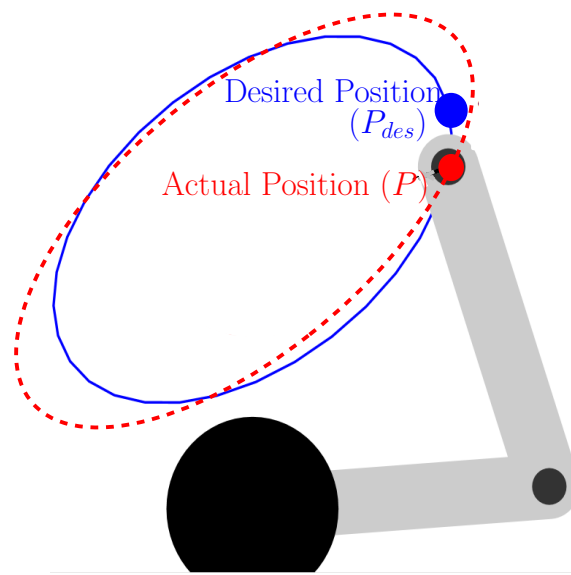}
\caption{Exercise environment for the advanced training. $P$ and $P_{des}$ are the current and desired robot end-effector position (red and blue dot respectively).}
\label{fig_GUI_Impedance}
\end{center}
\end{figure}

\noindent During the exercise protocol, the subject was required to follow the desired position while receiving feedback from his or her current position. Thus, a graphical user interface (GUI) is used to provide the required visual feedback. The complete environment can be seen in Figure \ref{fig_GUI_Impedance} where the blue line represents the ellipsoidal trajectory of fixed axis lengths and variable orientation that was designed based on the desired orientation using forward dynamics. The user's actual position $P$ (defined by the end-effector position of the robot) is labeled with a red dot. The desired position $P_{des}$ (rotating periodically over the blue ellipsoidal trajectory) is labeled with a blue dot. The user is required to track the blue dot as close as possible. EMG was used to measure and provide biofeedback of the muscle activations to the system. The model-free framework making use of the robot positions and muscle activations regulates the orientation aiming to maximize muscle performance. This muscle performance is defined as the moving average of the multiplication of a muscle weight vector and the muscle activation derived as follows:
\begin{equation}
\label{Eqn_Cost_SVO_Tr_1}
\max_{\theta} \quad J(t)= \frac{t_s}{t_{rev}}\sum^t_{(i=t-t_{rev}/t_s)}\bigg(W_mM(t)\bigg),
\end{equation} 
\noindent where $\theta$ is the ellipsoidal trajectory orientation, $t_s$ is the sample time set at 5e-4 seconds, $t_{rev}$ is the frequency of revolution of the reference cursor (blue dot), $t$ is the current time, $W_m$ is the weight muscle vector, and $M$ is the vector of muscle activations. The weight muscle vector receives the priority values for each muscle. This priority can be selected by professionals or therapists to emphasize or de-emphasize certain muscle groups. For this study, 3 possible priorities were selected to be assigned arbitrarily. These priorities were low, medium, and high related to priority gains of 1, 3, and 5 respectively. Therefore, muscles with higher priority (higher gain) would receive more maximization focus.

\section{Dynamics}\label{S_Dynamics}
\noindent The AEM used in this study was a WAM robot. The WAM is 4 DOFs articulated linkage robot from Barrett Advanced Robotics. This lightweight cable-driven manipulator is exceptionally dexterous, low-friction naturally backdrivable. Therefore, the robot is able to provide an almost negligible interaction resistance between the subject and the machine. The dynamics of this robot \cite{ROBOT_MODEL} given in joint coordinates are derived as:
\begin{equation}
\label{DE1}
D(q)\ddot{q}+C(q,\dot{q})\dot{q}+g(q)=\tau+J^TF_{ext},
\end{equation}
where $q=[q_1,q_2,q_3,q_4]$ is a vector of joint displacements, $D(q)$ is the inertia matrix, $C(q,\dot{q})$ is the centripetal and Coriolis effects, $g(q)$ is the gravity vector, $\tau$ is the control torque, $J$ is the Jacobian, and $F_{ext}$ is the external force representing the interaction force between subject and robot.

\noindent Muscle dynamics are mainly affected by time-invariant independent, organismic, and intervening variables such as robotic parameters, musculoskeletal distribution, performance status, level of hydration, and mood; and time-variant variables like muscle temperature and fatigue. As a result, muscle dynamics can be described as an unknown dynamic equation as follows:
\begin{equation}
\label{M_fcn}
 \dot M=f(M,u(t)),
\end{equation}
where $M=[M_1,M_2,M_3,M_4]$ is a vector of muscle activations and $u(t)$ is a nonlinear time-varying function including the ellipsoidal orientation ($\theta$) and all the unknown and unmeasurable variables.  

\section{Control Architecture}\label{S_Control}
\noindent Two controllers working simultaneously were involved in this project. The first was a PD-gravity compensator controller working isolated on the AEM. The second controller works on the complete HMI system relating the robot dynamics with muscle performance.\\
 
\subsection{Robot - PD-Gravity Compensation Control}
\noindent In order to work on the planar ellipsoidal trajectory, 2 of the 4 joints of the robot were fixed by applying a PD controller acting on the third and fourth joints. Gravity compensation was applied to increase robot maneuverability. The control law was defined as:
\begin{equation}
\tau=P_{gain}e+D_{gain}\dot e+g(q),
\end{equation}
where $P_{gain}$ and $D_{gain}$ are diagonal PD gain matrices (zero diagonal values for the first 2 DOFs), and $e$ is the error vector between the desired and the current joint positions. The end-effector robot position ($P$) is sent and showed in the GUI during the complete experiment. 
 
\subsection{Main Controller - Extremum Seeking Control (ESC)}
\begin{figure}[ht]
\begin{center}
\includegraphics[width=8.4cm]{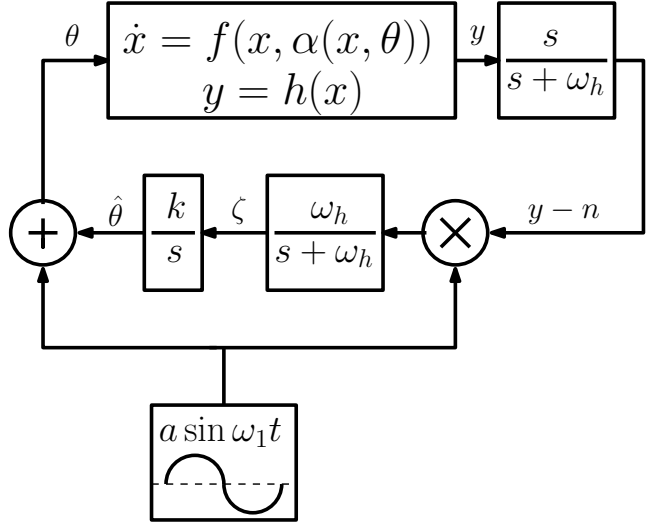}
\caption{Scheme of ESC for SISO with deterministic perturbation, where $\omega_l$ and $\omega_h$ are the low-pass and high-pass cutoff frequencies respectively, $k$ is the gain, $a$ is the amplitude of the perturbation, and $\omega_1$ is the frequency of the perturbation.}
\label{fig_ESC_scheme}
\end{center}
\end{figure}

\noindent The main controller is based on the use of the perturbation-based Extremum Seeking Control (ESC) \cite{ESC_1,ES_Book0}. The ESC scheme can be seen in Figure \ref{fig_ESC_scheme}. ESC is a branch of adaptive control developed for optimization. The aim of this control is to enforce the output of a dynamical system to converge to an unknown maximum or minimum operating points. Several ESC algorithms have been developed and studied from the first appearance in the 1920's \cite{ES_book1,ES_Book0}. Some of them are model-based methods. However, the popularity of ESC lies in its model-free optimization algorithms and its ability to operate in real-time.\\

\noindent Comparable to any other optimization method, the basis of the ESC is provided by the general optimal conditions \cite{Optimality}. For a general nonlinear system:
\begin{equation}
\dot{x} = f(t,x,u),
\end{equation}
\noindent and the following performance output:
\begin{equation}
y = J (t,\theta),
\end{equation}

\noindent where $t$ is the time, $x^T=[x_1,x_2,...,x_n]$ is the state vector, $u$ is the input, and the functions $f : D \times \mathbb{R} \to \mathbb{R} ^n$ and $J : D \to \mathbb{R}$ are sufficiently smooth on $D 	\subseteq \mathbb{R}^n$.\\

\noindent Assuming stability on the system (or being stabilizable) at each of the equilibrium points by a local feedback controller. Suppose the existence of the following smooth control law:
\begin{equation}
u=\alpha(x,\theta),
\end{equation}
\noindent parameterized by a scalar parameter $\theta$. The closed-loop system becomes:
\begin{equation}
\dot{x} = f\bigg(t,x,\alpha(x,\theta)\bigg),
\end{equation}
\noindent therefore, the equilibrium points of the system will be parameterized by $\theta$.\\

\noindent The existence of a local minimizer (or maximizer) is guaranteed under specific assumptions about the closed-loop system. Assumptions and stability analysis are derived in \cite{ES_book1,ES_book2}.\\

\noindent ESC was selected to deal with the nonlinear time-varying muscle dynamics including the unknown and unmeasurable parameters. This controller receives processed muscle performance. Then, based on its variation, the controller regulates the orientation of the ellipsoidal curve ($\theta$) to be transformed using forward dynamics into the desired trajectory ($P_{des}$).\\

\noindent The ESC parameters used on this project (related to Figure \ref{fig_ESC_scheme}) were 0.1, 1 $rad/s$, 0.1 $rad/s$ ,0.5 $rad/s$, and 1000 for the perturbation amplitude ($a$), perturbation frequency ($\omega_1$), low-pass filter cutoff frequency ($\omega_l$), high-pass filter cutoff frequency ($\omega_h$), and gain ($k$) respectively.

\subsection{Closed-loop system scheme for the model-free optimization}

\begin{figure}[ht]
\begin{center}
\includegraphics[width=8.7cm]{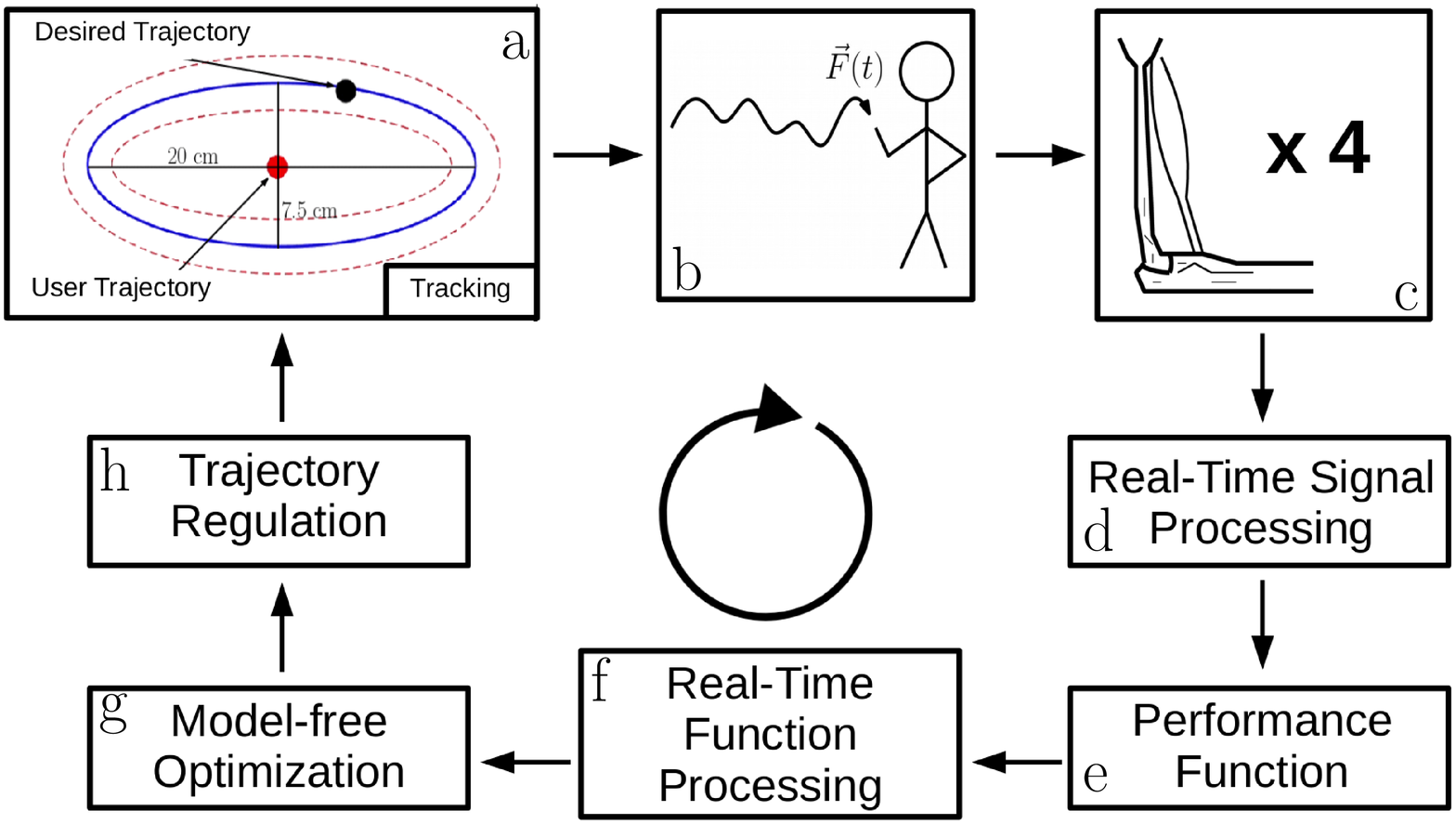}
\caption{Closed-loop system scheme for the model-free optimization approach.}
\label{fig_Scheme_2}
\end{center}
\end{figure}

\noindent The closed-loop system scheme can be seen in Figure \ref{fig_Scheme_2}. The process begins with the GUI providing visual feedback to the user (see Figure \ref{fig_Scheme_2}-a). Next, the user drives the end-effector of the robot following the desired ellipsoidal trajectory (see Figures \ref{fig_WAM_Exp} and \ref{fig_Scheme_2}-b). As a result of the training, muscle activations are produced and measured with EMGs as raw signals (see Figure \ref{fig_Scheme_2}-c). These signals are recorded at a frequency of 2 kHz and real-time processed to obtain muscular activation (see Figure \ref{fig_Scheme_2}-d). The complete process includes a bandpass second-order Butterworth filter between 30 and 950 Hz, a full-wave rectification, and a low pass second-order Butterworth filter at 50 Hz. The muscular activations are then used to compute the overall muscle performance which is then maximized (see Figure \ref{fig_Scheme_2}-e). Due to the high sensitivity of the muscle signal, a second-order filter is applied at the output of the performance function (see Figure \ref{fig_Scheme_2}-f). This filter introduces a delay to the signal suppressing possible perturbations. The main controller running the model-free optimization (see Figure \ref{fig_Scheme_2}-g) makes use of the processed muscle performance to regulate the orientation of the visual geometric path to be tracked by the subject (see Figure \ref{fig_Scheme_2}-h). While the orientation is being adjusted to achieve maximal performance, the cycle continues to repeat. The optimal solution is considered found after the optimization variable remains between $\pm 10^\circ$ for at least 10 seconds (convergence criterion).

\section{Experimental Protocol}\label{S_experimental_protocol}
\begin{figure}[ht]
\begin{center}
\includegraphics[width=8.5cm]{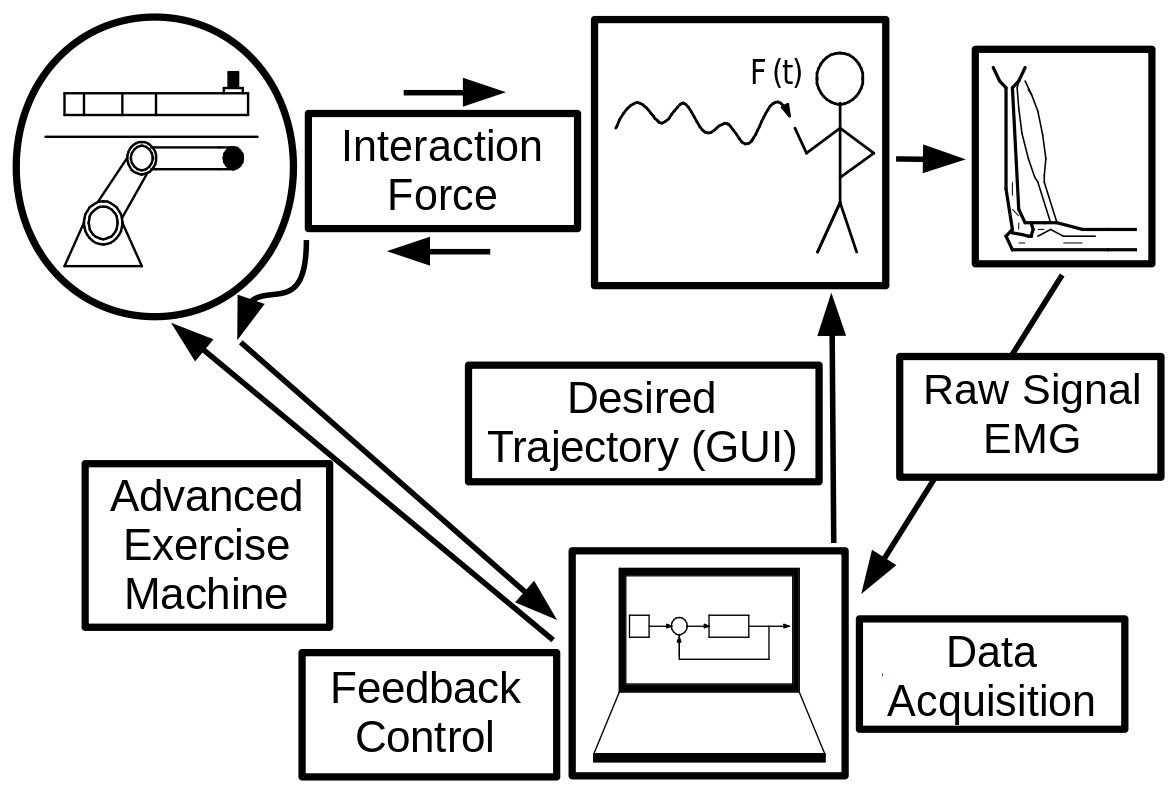}
\caption{Scheme of the HMI system for training optimization.}
\label{fig_Scheme_1}
\end{center}
\end{figure}

\noindent The complete HMI system used for the experiments can be seen in Figure \ref{fig_Scheme_1}. This system includes a WAM robot as the AEM, a desktop PC with Xenomai Kernel, a laptop PC with MATLAB and ControlDesk, a MicroLabBox real-time data acquisition and control system (dSPACE GmbH, Paderborn, Germany), a set of wireless EMG sensors (Trigno Wireless EMG, Delsys Inc.), and 2 DAQs (Labjack U3-HV and Labjack T7). The PC with Xenomai Kernel was used to control the WAM robot by CANbus communication and send the robot states (related to the human arm position) to the laptop PC. The laptop PC was used to manage the setting parameters of the MicroLabBox. The MicroLabBox was utilized to read the variable states, perform the real-time processing, record the data, and run the main control. The Delsys Trigno Wireless EMG system aided in collecting the EMG activity of the four muscles selected for the study. The Labjack U3-HV was used to transmit the robot states from one PC to the other. The Labjack T7 was used to transmit the desired trajectory from the main controller to a visualization screen.\\

\begin{figure}[ht]
\begin{center}
\includegraphics[width=8.0cm]{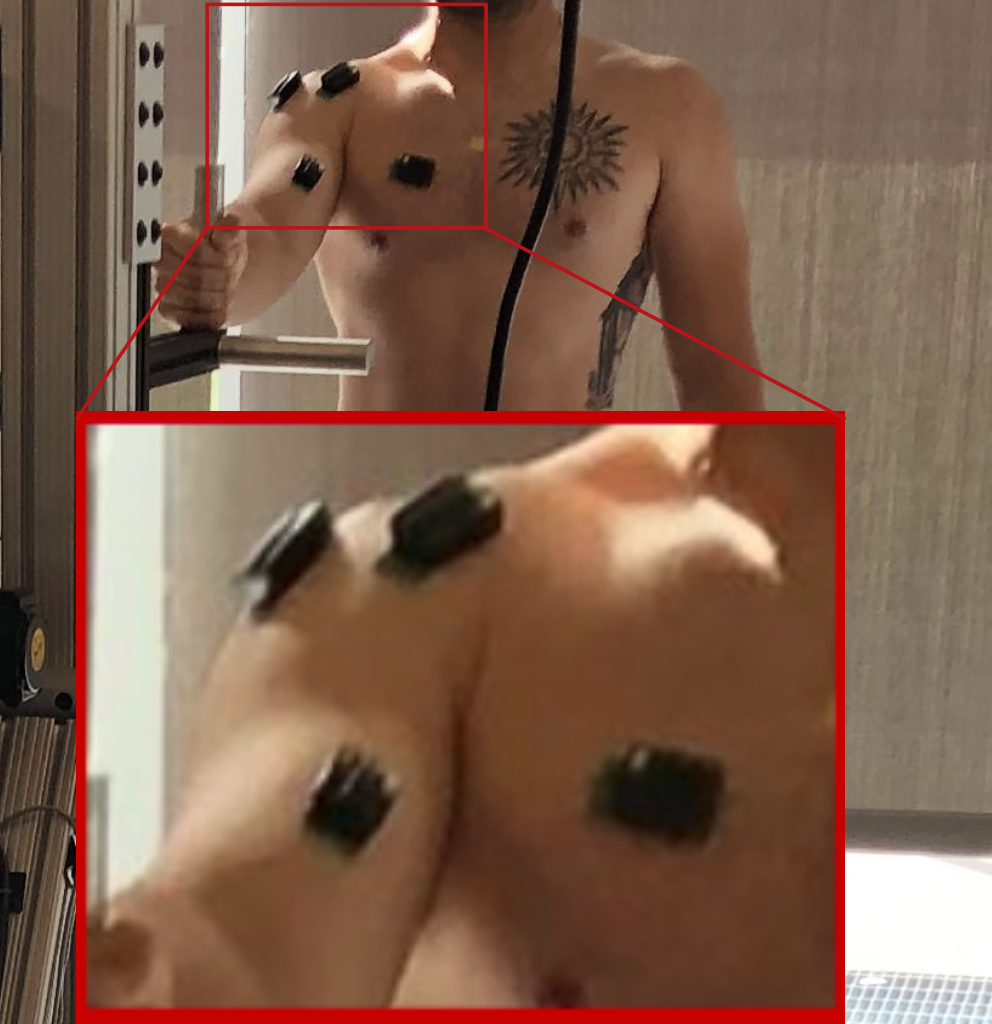}
\caption{4OptimX experiment configuration and EMG locations.}
\label{fig_4OptimxExp.png}
\end{center}
\end{figure}

Based on the experimental protocol, a subject was required to follow an ellipsoidal trajectory with variable orientation auto-regulated by using his muscular activations as biofeedback. Based on the type of movement and the range of motion, 4 muscles surrounding the glenohumeral joint \cite{Glenohumeral} (see Figure \ref{fig_4OptimxExp.png}) were selected in the following order:
\begin{enumerate}
\item \textbf{Lateral deltoid:} for shoulder abduction and stabilization during flexion.
\item \textbf{Anterior deltoid:} for shoulder flexion and inward rotation.
\item \textbf{Biceps brachii:} for lower arm flexion, and assistance and stabilization during shoulder flexion.
\item \textbf{Pectoralis major:} for shoulder adduction in the frontal plane.
\end{enumerate}

\begin{figure}[ht]
\begin{center}
\includegraphics[width=8.0cm]{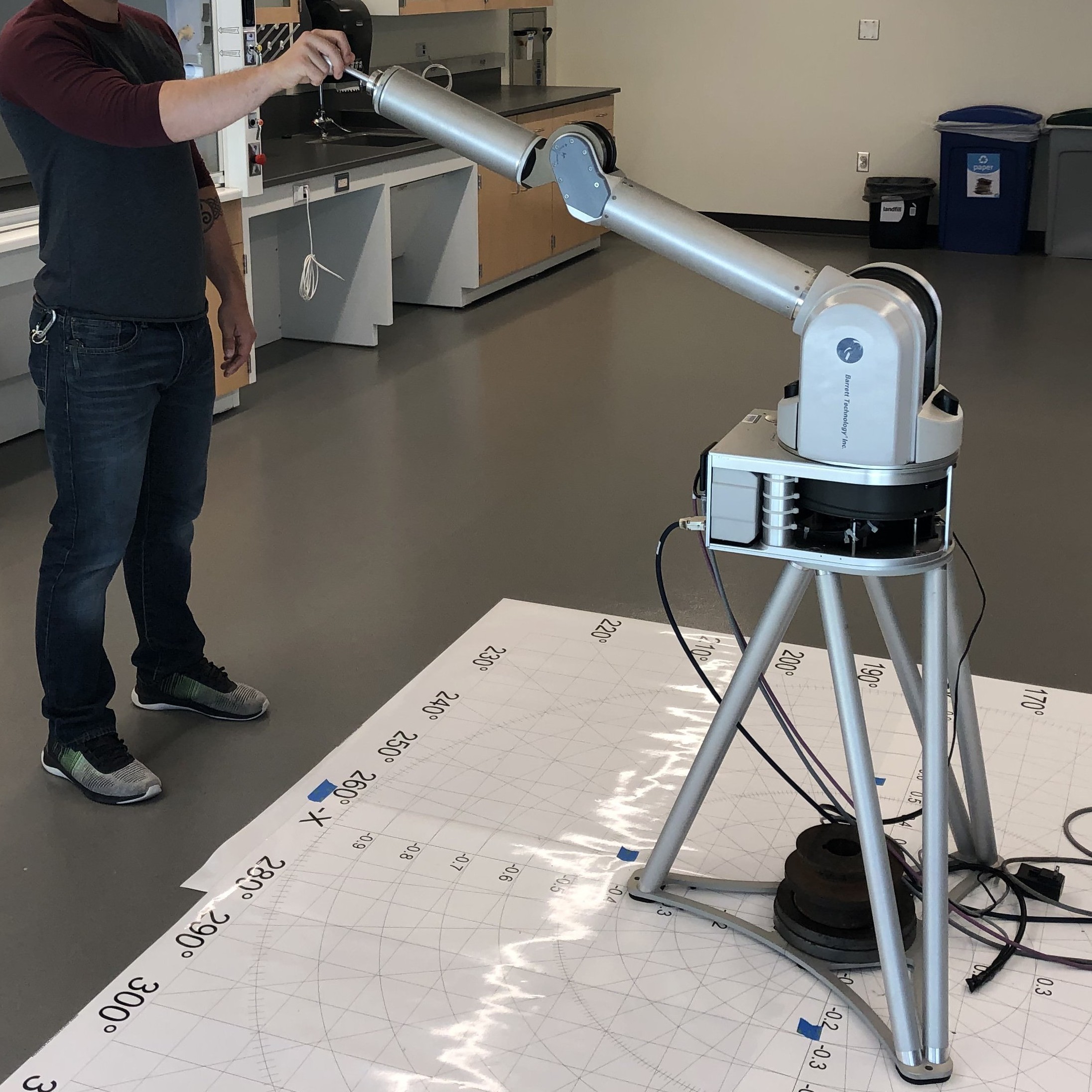}
\caption{Experiment of human training with AEM.}
\label{fig_WAM_Exp}
\end{center}
\end{figure}

\noindent One healthy right-handed male subject of the age of 23 years, height of 180.4 cm, and weight of 93.2 kg performed 3 sets of experiments on 3 different days (see Figure. \ref{fig_WAM_Exp}). The experimental procedure began with warm-up and isometric tests \citep{ISO2}. Isometric tests were used to assess muscle strength for the EMG sensor calibration. To establish these contractions, the subject was required to maintain different constant positions which muscles were capable of producing maximal muscular contracting forces (highest activations) \citep{ISO1}. Then, the experiment proceeded with the trials. During each day, one set of 2 trials (one trial next to the other one) was performed by using the same parameter configuration. The first set of experiments used the muscle weight vector $W_m=[1,5,3,5]$. This set gave the highest priority to the anterior deltoid and pectoralis major, medium priority to the biceps brachii, and the lowest priority to the lateral deltoid. The second set used the muscle weight vector $W_m=[3,5,1,1]$. This set gave the highest priority to the anterior deltoid, medium priority to the lateral deltoid, and the lowest to biceps brachii and pectoralis major. And the third set used the muscle weight vector $W_m=[1,1,5,5]$). This latest set gave the highest priority to the biceps brachii and pectoralis major and the lowest to the lateral and anterior deltoid.

\section{Results and Discussion}\label{S_Results}

\begin{figure}[ht]
\begin{center}
\includegraphics[width=9.0cm]{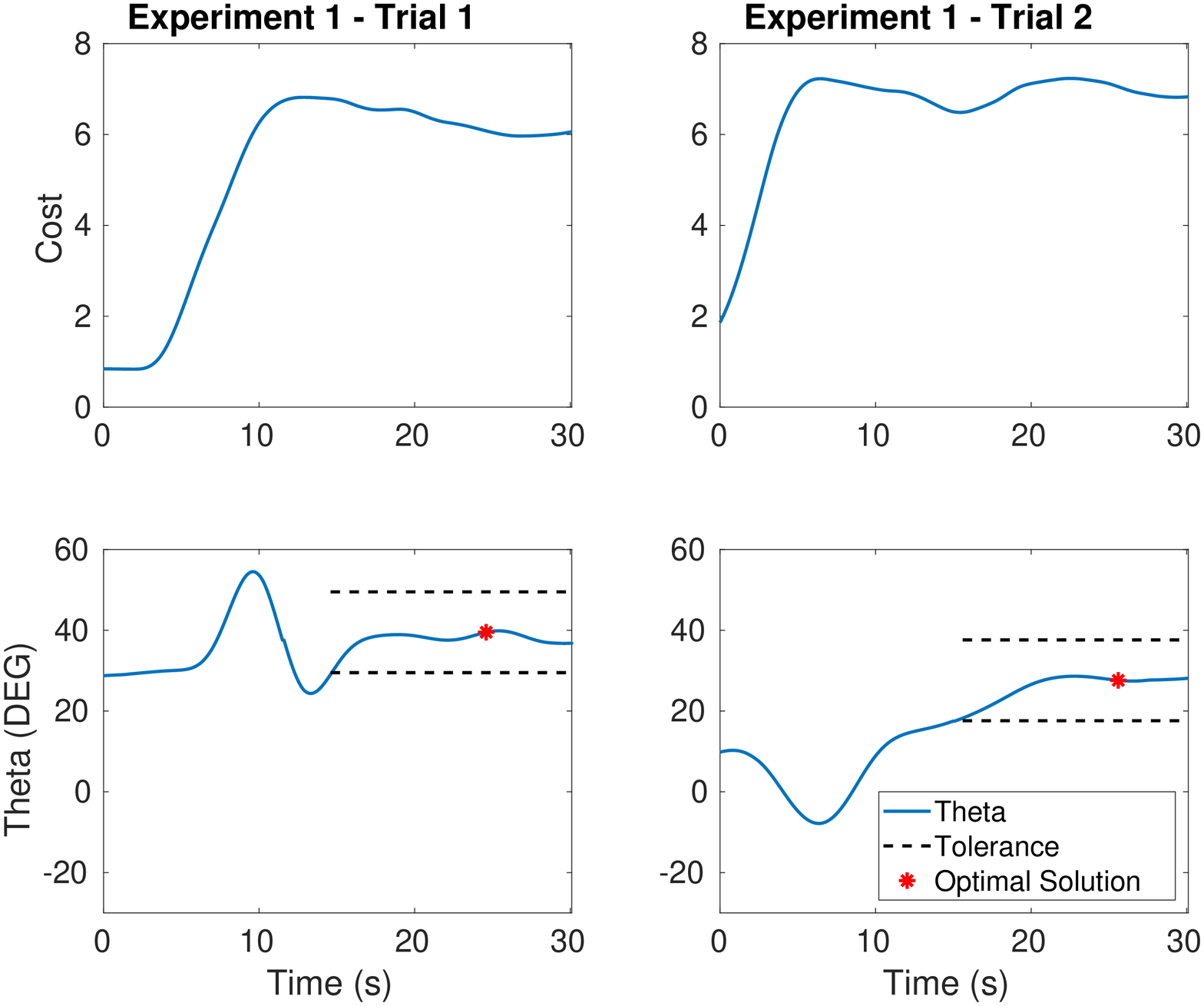}
\caption{First set of experiments with weight factor $W_m$=[1, 5, 3, 5]}
\label{fig_WAM_1}
\end{center}
\end{figure}

\begin{figure}[ht]
\begin{center}
\includegraphics[width=9.0cm]{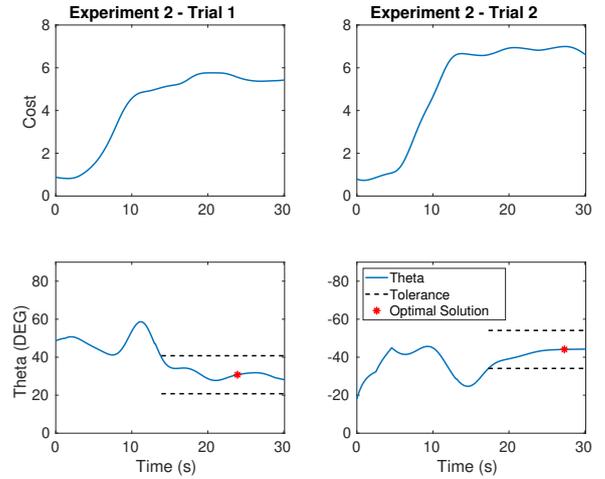}
\caption{Second set of experiments with weight factor $W_m$=[3, 5, 1, 1]}
\label{fig_WAM_2}
\end{center}
\end{figure}

\begin{figure}[ht]
\begin{center}
\includegraphics[width=9.0cm]{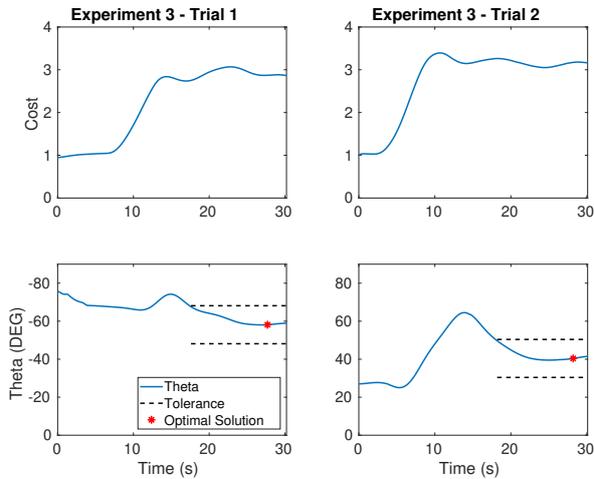}
\caption{Third set of experiments with weight factor $W_m$=[1, 1, 5, 5]}
\label{fig_WAM_3}
\end{center}
\end{figure}

\begin{table}[H]
\begin{center}
\caption{Experiments results - convergence of the solutions based on the selected weight factors (1, 3, and 5 representing the low, medium, and high priority gains respectively).}
\label{table_Results}
\begin{tabular}{c|c|c|c}

Weight factor &Trial&  Solution & Convergence \\
($W_m$)&(N$^\circ$)&  ($^\circ$)& Time (s) \\
\hline \\
$[1, 5, 3, 5]$&1 & 39.50 & 24.6\\
$[1, 5, 3, 5]$&2 & 27.60 &25.6\\
\\$
[3, 5,1,1]$& 1&30.77 & 23.9\\
$[3, 5,1,1]$& 2& -44.09 &27.3\\
\\
$[1,1,5,5]$& 1& -58.14 & 27.7\\
$[1,1,5,5]$& 2& 40.40 &28.2\\
\hline
\end{tabular}
\end{center}
\end{table}

\noindent Upon completion of the 6 experimental trials, differences in the muscle performance between trials from the same experimental set were revealed. A higher performance value was exhibited in the second trials during all the experimental sets. This result seems to be associated with the reduced effort capacity of the muscles to perform activity when they start to get fatigued. However, despite the difference in the muscle performance between trials, they converged to similar solutions depending on the initial conditions and the muscle weight vector. These results suggest that the formulated model-free optimization method successfully enabled the user to exercise optimally.\\ 

\begin{figure}[ht]
\begin{center}
\includegraphics[width=8cm]{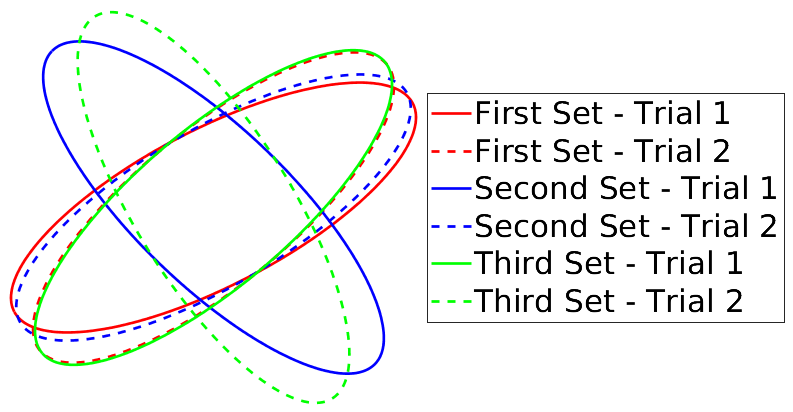}
\caption{Convergence solutions from the 3 sets of experiments.}
\label{fig_WAM_All_Results_All}
\end{center}
\end{figure}

\noindent It is important to consider that the optimal trajectory orientations don't only depend on the subject's musculoskeletal system (lengths, weights, distributions, etc), but also on his/her interaction with the machine. Although the robot offers little resistance, the user required to overcome the robot weight, inertia, and friction representing an additional resistance for the user and affecting the results. Besides, muscle dynamics are permanently changing. Therefore, ESC remains varying slightly even after reaching the optimal solution. However, based on the established convergence criterion, the solutions were computed and found in the neighborhood from $30^\circ$ to $40^\circ$ and its symmetric respect to the vertical axis from $-60^\circ$ to $-45^\circ$ (see Figure \ref{fig_WAM_All_Results_All}). These results suggest that an inclined ellipsoidal trajectory seems to provide a higher muscle demanding able to maximize the activations for training performance.

\noindent It is important to note that the success of the model-free approach lies in the accuracy of the user to track the desired trajectory and the selection of the configuration parameters. Regarding the accuracy of the tracking, it is recommended that the user is introduced and provided time to practice with the robot prior to testing so as to familiarize with the mechanisms. During the present study, tracking was acceptable and consistent between trials despite the presence of minor tracking errors. The mean of the RMS tracking errors was reported at 2.87 cm what proves the controller was robust enough to always converge to the neighborhoods of the optimizers. Regarding the configuration parameters, the framework might require some pre-tests to find a good combination of parameters for each subject. For instance, higher gains or frequencies might produce faster convergence, but at the cost of a higher sensitivity which is not recommended on this approach because of undesired performances. On the other hand, low gains or frequencies might never achieve a convergence or not being able to deal with the time-varying dynamics due to the fatigue and the thermogenic effect of the muscles. Nonetheless, the parameters chosen for the 3 experimental were accurate enough to produce robustness and a convenient convergence speed to deal with these variations. 

\section{Future Work}\label{S_Future}
\noindent Some challenges encountered while conducting these experiments include the high sensitivity of the EMGs. This proved problematic as the feedback resulted in involuntary elbow rotations, producing alterations in the muscle dynamics. The possibility of including human motion capture to feedback into the controller could be promising as a means of providing more information on concurrent motion dynamics. Another issue was associated with the electrical noise produced by the other systems being used in the testing environment. Aside from the current digital processes in this model, analog processing will be included as future development. In addition, as an alternate solution to muscle performance, estimation methods will be tested. Due to the high computing cost of the high-frequency sampling, estimation methods such as Kalman filtering show potential in improving the efficiency of this method.\\

\noindent Coincidentally in the 3 experimental sets, longer convergence times were observed in the second trials (after a few minutes of working out). This result suggests that there is a possible relationship between convergence time and fatigue. It is known that muscles consist of many motor units that are not fully active at the beginning of the workout, but they start to activate together with the increasing of fatigue \cite{Fatigue_1,Fatigue_2}. The increase in the activation of motor units produces an increase in the muscle activations similarly to an increasing in the framework gain producing a higher sensitivity, and thus undesired performances. These sudden changes might not only delay the convergence but also they might even block it. Therefore, it can be concluded that independently of the accuracy in the initial setting parameter selection, recalibration might become needed after a few minutes of training. The increase in the sensitivity previously observed might be solved by decreasing the framework gain, thus future studies could include automatic parameter calibrations to overcome this current limitation.\\

\noindent As a final conclusion, results suggest the feasibility of the formulated model-free optimization for trajectory orientation by targeting muscular performance. The feasibility of the approach will be also tested for impedance regulation. As a long term objective, research on multi-variable optimization of trajectory and impedance parameters using more physiological effects related to the cardiovascular and cardiorespiratory systems (such as heart rate, oxygen consumption, and energy expenditure) will be considered. This approach will make it possible to optimize the training of an individual to provide both an aerobic and anaerobic stimulus.\\

\section*{Acknowledgements}
\noindent We wish to confirm that there are no known conflicts of interest associated with this publication and there has been no significant financial support for this work that could have influenced its outcome. This research was supported by the NSF (grant 1544702). This research has an IRB provided by Cleveland State University (reference number 30305-RIC-HS) which covers for Ethical Approval. Subjects on this research have read and signed informed consents.

\bibliographystyle{elsarticle-num}
\bibliography{elsarticle-template}

\end{document}